\documentclass[5p, final]{elsarticle}
\usepackage{setspace}

\usepackage{amssymb}
\usepackage{amsthm}
\usepackage{amsmath}
\usepackage{subfigure}
\usepackage{hyperref}
\hypersetup{pdfauthor=author}
\usepackage{booktabs}
\usepackage{caption}
\usepackage{setspace}
\usepackage[dvipsnames]{xcolor}
\usepackage{mathtools}

\usepackage{url}

\journal{Neurocomputing}

\usepackage{amsmath}

\DeclareMathOperator*{\argmin}{arg\,min}

\usepackage[noend]{algorithm2e}

\begin{document}

\begin{frontmatter}
  
  \title{The Conditioning Bias in Binary Decision Trees and Random Forests and Its Elimination}

  \author[1]{G\'abor Tim\'ar}
  \ead{gtimar@ua.pt}
  \author[2]{Gy\"orgy Kov\'acs\corref{cor1}%
  }
  \ead{gyuriofkovacs@gmail.com}

  \cortext[cor1]{Corresponding author}

  \address[1]{Departamento de F\'isica da Universidade de Aveiro \& I3N, Campus Universit\'ario de Santiago, 3810-193 Aveiro, Portugal}
  \address[2]{Analytical Minds Ltd., \'Arp\'ad street 5, Beregsur\'any 4933, Hungary}

\begin{abstract}

Decision tree and random forest classification and regression are some of the most widely used in machine learning approaches. Binary decision tree implementations commonly use conditioning in the form 'feature $\leq$ (or $<$) threshold', with the threshold being the midpoint between two observed feature values. In this paper, we investigate the bias introduced by the choice of conditioning operator (an intrinsic property of implementations) in the presence of features with lattice characteristics. We propose techniques to eliminate this bias, requiring an additional prediction with decision trees and incurring no cost for random forests.
Using 20 classification and 20 regression datasets, we demonstrate that the bias can lead to statistically significant differences in terms of AUC and $r^2$ scores. The proposed techniques successfully mitigate the bias, compared to the worst-case scenario, statistically significant improvements of up to 0.1-0.2 percentage points of AUC and $r^2$ scores were achieved and the improvement of 1.5 percentage points of $r^2$ score was measured in the most sensitive case of random forest regression.
The implementation of the study is available on GitHub at the following repository: \url{https://github.com/gykovacs/conditioning_bias}.

\end{abstract}



\end{frontmatter}


\thispagestyle{empty}
\section{Introduction}

Decision trees for supervised machine learning problems were proposed more than 40 years ago \cite{chaid-earlier, id3}, and they are still preferred techniques when the interpretability of predictive models is a requirement, especially in fields such as medical \cite{dt-medicine-000, dt-medicine-001, dt-medicine-002, dt-medicine-003, dt-medicine-004} and financial \cite{dt-business-000, dt-business-002, dt-business-003} decision-making. Derivative techniques, like various types of tree ensembles (random forests \cite{rf0}, boosting \cite{xgboost, lightgbm}), are among the best-performing general-purpose classifiers and regressors, consistently outperforming other techniques in various data science problems and competitions \cite{higgs, load-forecasting}.

\begin{figure}
  \begin{center}
    \includegraphics[width=0.47\textwidth]{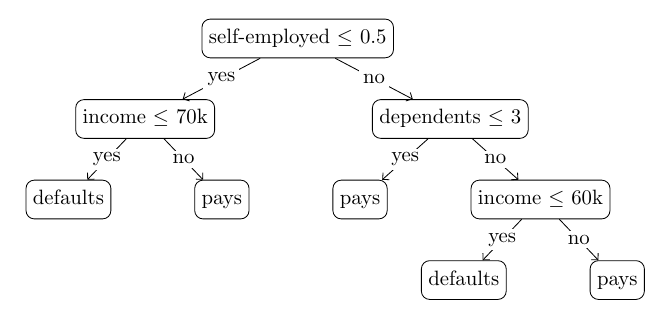}
  \end{center}
  \caption{A decision tree illustrating a loan approval problem, based on three features (self-employment (0 or 1), annual income, and the number of dependents). The nodes without conditions contain the prediction labels: whether the applicant would default or pay back the loan. For the underlying sample data, see \ref{sec:sample}.}
  \label{fig:dt}
\end{figure}

Decision trees are directed, acyclic graphs in which internal nodes represent tests related to the features of a problem. The branches correspond to the outcomes of these tests, and the leaf nodes contain estimates characterizing the partition of the feature space represented by the conditions along the path from the root to the leaf node. Leaf nodes typically hold class labels or probabilities for classification and continuous point estimates for regression problems. During inference, the observation vector is routed to a leaf node, and the contents of the leaf node are assigned to it as predictions (see Figure \ref{fig:dt} for an illustration).

It was shown early that finding the optimal decision tree for a given dataset is an NP-complete problem \cite{dt-np-complete}. Although numerous \emph{global optimization} \cite{evolutionary-1} and \emph{bottom-up} heuristics \cite{dt-butia, dt-bottom-up} have been proposed for tree induction, \emph{top-down} greedy techniques remain the most popular \cite{dt-partitioning, dt-induction}.

These approaches recursively partition the feature space by imposing equality conditions on categorical attributes or thresholding conditions on numerical attributes of the feature vectors. The conditions are optimized to decrease the impurity of target labels in the resulting partitions. In classification problems, impurity is typically measured in terms of \emph{entropy} or \emph{Gini-score} \cite{treebook}, while in regression problems, the primary impurity measure is usually the variance \cite{treebook}. Two variations of this scheme, C4.5 and \emph{Classification and Regression Trees} (CART), are considered two of the most important algorithms in data mining \cite{top10}, with implementations available in various open-source (\verb|sklearn| \cite{sklearn}, \verb|tree| \cite{tree}, WEKA \cite{weka}) packages and commercial software (Matlab \cite{matlab}, Wolfram Mathematica \cite{mathematica}). Additionally, CART trees have proven to be efficient weak learners in highly effective ensemble techniques like \emph{random forests} \cite{rf0}, \emph{eXtreme Gradient Boosting machine} (xgboost) \cite{xgboost}, and \emph{Light Gradient Boosting Machine} (lightgbm) \cite{lightgbm}.

Although the concept of supervised decision trees is well-established, ongoing research in the field continues to introduce new ideas. Recently, impurity measures that address certain biases of conventional ones have been proposed. Examples include mutual information \cite{dt-entropy-split-mutual-information}, averaged information gain \cite{dt-average-gain}, minimum entropy of error \cite{dt-gain-minimum-entropy-of-error}, permutation statistics \cite{dt-split-others-permutation}, fuzzy rules \cite{dt-split-fuzzy}, the use of rough sets \cite{dt-split-rough-set}, and the comparison of distributions \cite{dt-chi2-0, dt-chi2-1, dt-split-ks, ds-split-ks1}. In contrast to univariate (axis-parallel) splitting functions, techniques like \cite{dt-split-oblique, dt-split-oblique-1, oc1, hhcart} operate with multivariate splits, and the use of Support Vector Machine decision boundary for multivariate splitting has also been proposed \cite{dt-split-oblique-svm}.
Regarding regularization, various pruning techniques \cite{dt-pruning-evolutionary, dt-pruning-max-heap} have been introduced to simplify the structure of already induced decision trees. It is worth mentioning that tree learning has been adapted for tasks beyond the classic classification and regression use cases in machine learning, such as survival rate prediction \cite{survival-forest} and density estimation \cite{density-tree}.
Finally, in line with recent trends in machine learning, research has delved into tree learning with privacy \cite{privacy}, adaptation to big data \cite{bigdata}, and the concept of deep models \cite{decision-stream}. For further details on the key concepts of decision tree learning and tree-based machine learning, we refer to \cite{treebook, survey, tutorial}.

Despite the numerous variants of decision tree learning proposed over the last decades, it is noteworthy that many applications, including those cited earlier, rely on robust implementations of binary CART decision trees (and their ensembles) \cite{sklearn, tree, caret, xgboost, lightgbm}. These tree induction techniques usually assume that all features are continuous or are embedded into a continuous representation, hence, the partitioning of the training data and the feature space by thresholding conditions (such as the ones in Figure \ref{fig:dt}) is applicable along all axes. However, in general, the partitioning of the finite amount of training data does not imply a unique partitioning of the continuous feature space. One ambiguity arises from the fact that if there are no observations for a feature between the observed $x^l$ and $x^u$ values, any threshold $t\in]x^l, x^u[$ leads to the same partitioning of training data. This ambiguity is usually resolved in alignment with the assumption of the features being continuous: if the threshold should fall between the observed $x^l$ and $x^u$ values, the mid-point $t=(x^l + x^u)/2$ is used, thus, assigning 50\% of the unlabeled interval to both the left and right partitions. Another ambiguity arises from the choice of conditioning: the Python implementation \emph{sklearn} \cite{sklearn} uses the operator $\leq$ in the form $x \leq t$, leading to partitions open from the left and closed from the right along each axis (as in Figure \ref{fig:dt}). In contrast, the R implementation \emph{tree} \cite{tree} uses conditions of the form $x < t$ leading to partitions closed from the left and open from the right along all axes. Considering the presumption of continuous features, the choice of conditioning operator should have a marginal effect, as a feature takes the exact value of a threshold with negligible probability. However, when the domain of a feature has lattice characteristics, the thresholds determined as the mid-points of observations can be aligned with the feature values making the routing dependent on the choice of conditioning operator. We refer to this dependence of predictions from the conditioning (which is an intrinsic property of popular implementations) as the \emph{conditioning bias}.

Although it might seem to be a marginal effect, due to the popularity of binary decision trees and random forests, we devote this paper to the investigation of the conditioning bias. 
The contributions of the paper to the field can be summarized as follows:
\begin{enumerate}
  \item We demonstrate and establish, with statistical significance, the emergence of the conditioning bias in binary decision trees and related techniques when dealing with features exhibiting lattice characteristics.
  \item Depending on the accessibility of the internals of fitted trees, various techniques are proposed to reduce the bias. 
  \item Through extensive experiments we demonstrate that the proposed methods can lead to statistically significant improvements of up to 0.1-0.2 percentage points in AUC and $r^2$ scores, but 1.5 percentage point of $r^2$ improvement was also observed in the most sensitive case.
  \item As an important finding, the elimination of the bias in random forests incurs no additional computational costs, neither in training nor in inference time. The proposed method is a \emph{free-lunch} improvement for applications using random forests on datasets with features exhibiting lattice characteristics.
\end{enumerate}

The proposed techniques and the completely reproducible implementation of the study are shared with the community in the GitHub repository \url{https://github.com/gykovacs/conditioning_bias}.

The paper is organized as follows: In Section \ref{sec:problem}, we formulate the problem and demonstrate its existence qualitatively. Section \ref{sec:proposed} describes the proposed solutions. The experimental setup and the results of the tests are presented and discussed in Section \ref{sec:tests}, and finally, conclusions are drawn in Section \ref{sec:summary}.

\section{Problem Formulation}
\label{sec:problem}

In this section, we provide an introduction to decision tree learning (subsections \ref{sec:problem-intro} and \ref{sec:problem-split}) and a detailed formulation of the problem we address (subsection \ref{sec:problem-bias}). 

In the rest of the paper, normal, boldface and calligraphic typesetting are used to denote scalars ($d \in\mathbb{Z}$), vectors ($\mathbf{x}\in\mathbb{Z}^d$) and sets ($\mathcal{S}\subset\mathbb{Z}$), respectively. Elements of sequences and sets are indexed by superscripts, coordinates of vectors are accessed by subscripts.

\subsection{A brief introduction to binary decision tree induction and inference}
\label{sec:problem-intro}

The popular CART-type \cite{cart} binary decision trees \cite{sklearn, tree, caret} \emph{assume all features being continuous}, and build the decision trees by recursively partitioning the training set to improve the purity of training labels within the partitions by the following steps. Let $(\mathbf{x}^i, y^i)_{i=1}^{N}, \mathbf{x}^i \in \mathbb{R}^{d}, y^i\in\mathbb{R}$ denote the feature vectors and corresponding labels at a particular node (initially the entire training set at the root node):
\begin{enumerate}
\item If the impurity of labels $I((y^i)_{i=1}^{N})$ is below a threshold, the node becomes a leaf node and the relevant statistics of the labels are recorded for prediction. 
\item Otherwise, the best feature $f^* \in \lbrace 1, \dots, d\rbrace$ and threshold $t^* \in \mathbb{R}$ are determined  
such that partitioning the training set into left $\mathcal{L} = \lbrace j \vert  \mathbf{x}^j_{f^*} \leq t^*\rbrace$ and right $\mathcal{R} = \lbrace j \vert  \mathbf{x}^j_{f^*} > t^*\rbrace$ subsets minimizes the total impurity $I((y^j)_{j\in\mathcal{L}}) + I((y^j)_{j\in\mathcal{R}})$. The parameters of the condition ($f^*$, $t^*$) are recorded and left and right child nodes are added by applying step 1 to the training records indexed by $\mathcal{L}$ and $\mathcal{R}$, respectively.
\end{enumerate}
Inference for a vector $\mathbf{x}$ is carried out by starting from the root node and recursively applying the $\mathbf{x}_{f} \leq t$ condition with the parameters $f$ and $t$ recorded in the actual node and routing $\mathbf{x}$ to the subtree corresponding to the output of the condition. When a leaf node is encountered, the recorded statistics (such as the class distribution for classification) are returned as the prediction for $\mathbf{x}$.

The use of partitions closed from the left by $\mathbf{x}_f\leq t$ is aligned with the Python implementation \emph{sklearn} \cite{sklearn}
For more details on the induction of decision trees and the hyperparameters controlling their complexity see \cite{treebook}.

\subsection{The splitting point and the conditioning operator}
\label{sec:problem-split}

\begin{figure}[t]
  \begin{center}
    \subfigure[\label{fig:split-cont-a}]{
      \includegraphics[width=0.47\textwidth]{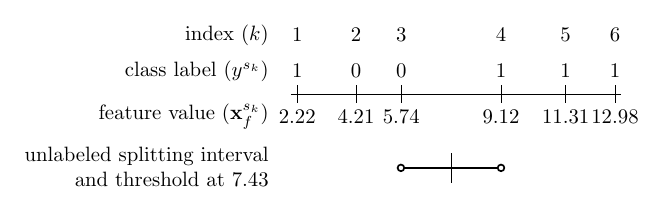}}
    \subfigure[\label{fig:split-cont-b}]{
      \includegraphics[width=0.47\textwidth]{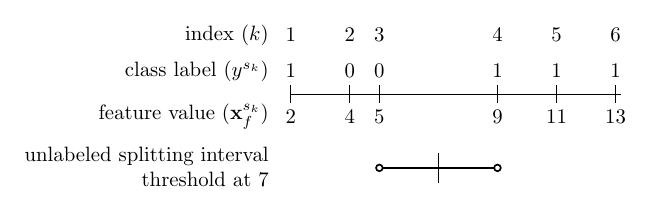}
    }
  \end{center}
  \caption{Finding the optimal partitioning in a classification problem for a continuous (a) and discrete (b) feature. In both cases, $k^* = 3$ leads to the lowest Gini-impurity for the resulting partitions (0.222). 
}

  \label{fig:split-cont}
\end{figure}

The key operation at a node is determining the optimal threshold $t$ for a feature $f$, such that conditioning the training records by $\mathbf{x}_f \leq t$, the highest reduction of impurity is gained in the resulting partitions. 
In practice, the feature values are sorted $\mathbf{x}^{s_1}_f \leq \dots \leq \mathbf{x}^{s_N}_f$, and the index of the rightmost value in the left partition is determined as
\begin{equation}
k^* = \argmin\limits_{\substack{k \in \lbrace 1, \dots, N - 1 \rbrace\\ \mathbf{x}^{s_{k}} \neq \mathbf{x}^{s_{k + 1}}}}I((y^{s_i})_{i=1}^{k}) + I((y^{s_i})_{i=k+1}^{N}).
\end{equation}
The optimization determines that the splitting needs to take place between $x^{s_{k^*}}_f$ and $x^{s_{k^* + 1}}_f$, but there is a freedom in choosing any threshold from this interval. Given that there are no observations in the interval, and in line with the assumption that the feature is continuous, most implementations choose the mid-point $t = (x^{s_{k^*}}_f + x^{s_{k^* + 1}}_f)/2$ as threshold, assigning 50\% of the splitting interval to both partitions (see Figure \ref{fig:split-cont}). We also highlight that using the mid-points as thresholds, \emph{the choice of conditioning ($<$ or $\leq$) does not affect the tree induction}: the same tree with the same thresholds is induced with both conditioning operators.

\subsection{The conditioning bias}
\label{sec:problem-bias}

With practically continuous features, the choice of conditioning ($<$ or $\leq$) has a marginal effect on inference since observations are unlikely to take the exact value of any threshold. However, if the domain of a feature consists of a small number of equidistant values, the thresholds are more likely to coincide with observations, making the choice of conditioning have a non-negligible effect.

For example, the tree in Figure \ref{fig:dt} is induced from data with records where self-employed applicants have 2 or 4 dependents (see \ref{sec:sample}). Note, that the continuous filtering of training data during tree induction can naturally lead to this kind of underrepresentation of features beyond certain depths. The split on the attribute \emph{dependents} is to be made in the unlabeled interval $]2, 4[$. By thresholding at 3, the records to be inferred with 3 dependents would be routed according to the choice of conditioning: using $\leq$ to the left and using $<$ to the right.

The differing predictions achieved by differing conditioning operators cannot be unbiased at the same time. We refer to the potential bias caused by the data-independent choice of conditioning as the \emph{conditioning bias}.  
One can argue that the bias is due to using the mid-points as thresholds. However, the mid-points are reasonable choices for continuous features, any other choice of threshold would lead to asymmetries and consequent biases. Therefore, we consider the phenomenon to be an artifact of dealing with discrete features in a continuous framework, and attribute the bias to the choice of conditioning operator, which is an intrinsic property of popular implementations. 

The phenomenon is not limited to integer features or small data. The conditioning bias is likely to be present when (a) \emph{a feature takes values on a small lattice with high probability}, and (b) \emph{relatively deep trees are built} leading to underrepresented features beyond certain depths. Examples of features potentially affected by the bias include the \emph{age} attribute in medical datasets; \emph{rounded decimals} when features are observed to a certain number of digits; and \emph{monetary values} quoted in units of millions in financial datasets. 

In the rest of the paper, we address the mitigation of the conditioning bias from the theoretical and practical points of view.

\section{The Proposed Method}
\label{sec:proposed}

In this section, we consider various approaches to carry out predictions with the non-default conditioning operator of an implementation (subsection \ref{sec:equiv}); discuss the possibility of treating the conditioning as a hyperparameter (subsection \ref{sec:hyper}) and propose methods to mitigate the conditioning bias (subsection \ref{sec:integrate}). 

We highlight that the goal is to develop practical solutions that are applicable to the popular decision tree implementations (such as \cite{sklearn, tree, caret}), and require as little additional computation and coding as possible.

\subsection{Alternatives to changing the conditioning operator}
\label{sec:equiv}

In the methods we propose, it is crucial to carry out predictions with both conditioning operators. However, in the most popular decision tree implementations, the conditioning operator is an intrinsic property that cannot be changed. Since decision tree induction needs to be efficient and usually involves various advanced operations like pruning \cite{sklearn}, creating custom implementations where the conditioning operator is a parameter is impractical. This is especially true considering that the choice of conditioning does not affect the tree induction process. In this subsection, we discuss various ways prediction with the non-default operator of an implementation can be carried out, depending on the accessible details of trained decision trees.

\begin{figure*}
  \begin{center}
    \subfigure[\label{fig:dt2-a}]{\includegraphics[width=0.47\textwidth]{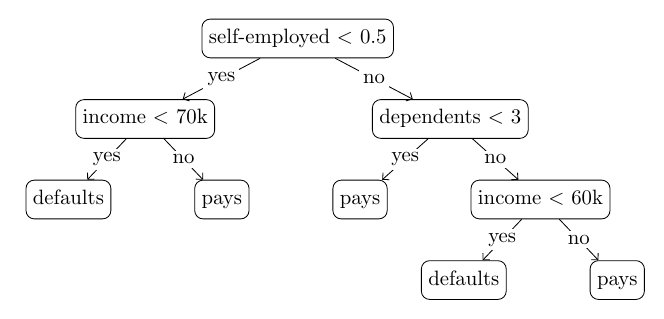}}
    \subfigure[\label{fig:dt2-b}]{\includegraphics[width=0.47\textwidth]{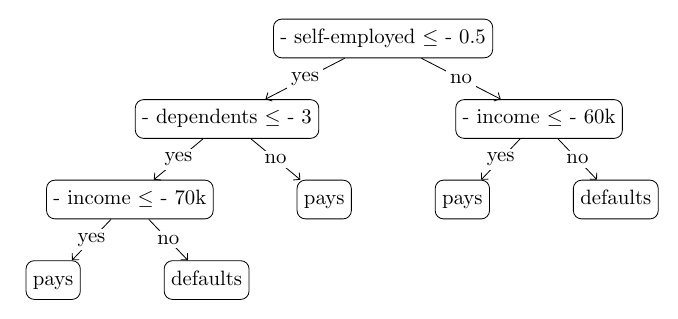}}
  \end{center}
  \caption{The decision tree in Figure \ref{fig:dt} with replaced conditioning operators (a), and its mirrored variant (b). One can check that given any feature vector $\mathbf{x}$, the prediction of $\mathbf{x}$ with (a) and that of $-\mathbf{x}$ with (b) leads to the same result along the same (but mirrored) decision path.}
  \label{fig:dt2}
\end{figure*}

Let $\mathcal{T} = (\mathbf{x}_i, y_i)_{i=1}^N$ denote a training set of records, and $T_{\mathcal{T}}$ be a decision tree fitted to $\mathcal{T}$. Without the loss of generality, we can suppose that the available tree induction and prediction implementation intrinsically uses the conditioning operator $\leq$. 
Let the prediction for $\mathbf{x}$ be denoted by $T_{\mathcal{T}}(\mathbf{x}, \leq)$. Our interest lies in inference with the same tree but using the non-default operator denoted as $T_{\mathcal{T}}(\mathbf{x}, <)$. As a practical example, we want to carry out inference with the tree in Figure \ref{fig:dt2-a}, given the possibility of inference with the tree in Figure \ref{fig:dt}.

The proposed workarounds to obtain $T_{\mathcal{T}}(\mathbf{x}, <)$ are based on the identity that the conditioning $\mathbf{x}_f < t$ is equivalent to $-\mathbf{x}_f \leq -t$ in the sense that if a vector $\mathbf{x}$ is routed to the left by $\mathbf{x}_f < t$, then $-\mathbf{x}$ is routed to the right by $-\mathbf{x}_f \leq -t$. Based on this identity, we introduce the operation of \emph{tree mirroring}, denoted by $\mathcal{M}[T_{\mathcal{T}}]$: interchanging the left and right subtrees in each node and replacing the threshold $t$ in each node by $-t$. 
The benefit of tree mirroring is that predicting $-\mathbf{x}$ with the mirrored tree leads to the same prediction along the same (but mirrored) decision path as predicting $\mathbf{x}$ by replacing the conditioning operator in the original tree. Formally, $T_{\mathcal{T}}(\mathbf{x}, <) = \mathcal{M}[T_{\mathcal{T}}](-\mathbf{x}, \leq)$.
For illustration, in Figure \ref{fig:dt2-b} we have plotted the mirrored version of the tree depicted in Figure \ref{fig:dt}. Considering a vector with a feature value coinciding with a threshold $\mathbf{x} = (\text{self-employment}: 0, \text{dependents}: 3, \text{income}: 50k)$, the predictions with the original tree (Figure \ref{fig:dt}) and the tree with the non-default operator (Figure \ref{fig:dt2-a}) lead to different results, however, the predictions using the operator $<$ (Figure \ref{fig:dt2-a}) and predicting $-\mathbf{x}$ with the mirrored tree (Figure \ref{fig:dt2-b}) are equivalent. 

Based on the equivalence of mirroring and operator change and depending on the accessibility of the fitted tree structure, we identified three ways to obtain $T_{\mathcal{T}}(\mathbf{x}, <)$:
\begin{enumerate}
\item If \emph{the tree structure is accessible and unmuteable}, one can implement the recursive tree inference algorithm discussed in subsection \ref{sec:problem-intro} \cite{treebook} using the operator $<$ for inference;
\item If \emph{the tree structure is accessible and mutable}, one can mirror the fitted tree structure and infer $-\mathbf{x}$. 
\item If \emph{the tree structure is not accessible} the mirrored tree can be induced directly from the opposites of feature vectors $\mathcal{T}' = (-\mathbf{x}_i, y_i)_{i=1}^N$. (We note, that depending on the implementation, in the case of multiple, equally good partitionings at a certain node, the tie might be resolved randomly, leading to differences in the tree structure.)
\end{enumerate}

\subsection{Treating the conditioning as a hyperparameter}
\label{sec:hyper}

From the general perspective, the conditioning operator could be treated as another hyperparameter of tree induction, similar to the maximum depth. 
Consequently, one can evaluate both operators (using the approaches proposed in subsection \ref{sec:equiv}) in a cross-validation scheme as part of the model selection and choose the one providing the highest performance in a particular problem. Moreover, one could argue that the conditioning should be specific to features, and all combinations of conditioning operators with lattice features should be optimized.

However, this approach has multiple drawbacks. Unlike other regularization parameters that serve as interpretable controls of smoothness, the choice of conditioning affects very specific edge cases and has a hardly predictable effect. This leads us to a major problem: the choice of conditioning cannot be cross-validated easily. In an ordinary cross-validation scenario, a certain percentage, say $P\%$, of the available data, is selected for training, and the remaining $(100-P)\%$ is used for testing (potentially in an iterated manner). If there is a difference in the prediction of the test data by the two operators, it is due to one of the test data feature values interfering with one of the thresholds. However, once the model is fit to the entire dataset, this interference is certainly eliminated, and other thresholds aligned on other feature values might have unpredictable effects on future, unseen data.

Due to the increased possibility of overfitting the selection of the optimal conditioning to the artifacts induced by cross-validation, we turn our attention to eliminating the downside effect by integrating out the effect of the conditioning operator.

\subsection{Integrating out the conditioning}
\label{sec:integrate}

Although one cannot know which conditioning leads to better results when a tree is trained with all available data, it is still reasonable to expect that the conditioning affects the predictions, and one operator could lead to better results than the other. We emphasize that in the form of the default operator, a choice is already made in popular implementations, and there is no reason to assume that there is anything other than a 50\% chance that the default operator does not cause a detrimental bias.

Motivated by the case of truly continuous features where, in principle, the choice of conditioning should have a close to 0 probability effect, a conservative yet reasonable approach to mitigate the bias is to integrate out the effect of the conditioning. Particularly, carrying out inference with both operators and taking the average of predictions clearly removes the dependence on the operator and can be expected to provide at least as good, but probably better predictions as the worst-performing conditioning.

In single decision trees, this can be done by using any of the approaches discussed in subsection \ref{sec:equiv} to carry out prediction with the non-default conditioning and averaging the results. In practice, the doubling of prediction time should not be a concern, since in most cases (a) decision trees are primarily used for their interpretability, and (b) inference with decision trees is relatively fast compared to many other ML techniques, such as k Nearest Neighbors. Interpretability is only affected when the predictions obtained by the two operators differ, and the averaging of two different predictions is still explainable by the ambiguous interference of a feature value and a condition.

In the case of ensembles, like random forests, one can follow the same approach and evaluate each tree of the ensemble with both operators. However, given the typically large number of estimators $N_e$, evaluating half of the trees with the default operator and the other half with the non-default conditioning should be a reasonably accurate approximation. Moreover, even if the tree induction implementation does not provide easy access to the internals, one can train one forest of $N_e/2$ trees on the original dataset and another forest of $N_e/2$ trees on the opposites (additive inverses) of feature vectors and use the average of the predictions for inference. Consequently, the approach can improve the quality of predictions with any random forest implementation for no additional cost either in training or prediction time apart from multiplying feature vectors by $-1$ and averaging two predictions.

\section{Tests and Results}
\label{sec:tests}

In this section, we present the datasets selected for evaluation (subsection \ref{sec:tests-datasets}), revisit the role of cross-validation (subsection \ref{sec:tests-cv}) and discuss the evaluation methodology (subsections \ref{sec:tests-model}, \ref{sec:tests-testing}), illustrate the bias we investigate quantitatively (subsection \ref{sec:tests-quant}), and finally, evaluate the proposed methods (subsection \ref{sec:tests-proposed}) using decision trees and random forests.

\subsection{Datasets}
\label{sec:tests-datasets}

To facilitate the selection of datasets for testing and evaluation, we introduce a heuristic condition. We term an attribute a \emph{lattice feature} and consider it susceptible to conditioning bias if there exists an observed domain value $\mathbf{x}_f$ and a positive real number $r\in\mathbb{R}^+$, such that both $\mathbf{x}_f-r$ and $\mathbf{x}_f+r$ are observed. The condition ensures that at least one threshold might align with a domain value. It is worth noting that this is not a necessary condition; previously unseen records can still interfere with thresholds, even if none of the features in the training data meet the condition.

We selected 20 binary classification and 20 regression datasets primarily from the UCI \cite{uci} and KEEL \cite{keel} repositories, using the criterion that each dataset should contain \emph{at least one lattice feature}. The dataset collection preceded all experiments and was not influenced by any experimental results. As observed in Table \ref{tab:datasets}, the datasets cover approximately an order of magnitude range in both the number of training records and the number of features. Interestingly, lattice features are more frequent in commonly used evaluation datasets than we anticipated; in many cases, all features meet the condition of being a lattice feature, despite the requirement being to have at least one such feature. To further support the claim regarding the frequency of lattice features, it is noteworthy that the average proportion of lattice features within the toy datasets available in the \emph{sklearn} \cite{sklearn} package is 97\% - almost all features of all toy datasets belong to this category.

The same preprocessing steps were applied to all datasets: first, missing data imputation with the most frequent value was carried out; then, one-hot encoding was applied to category features with at most 5 categories, and integer encoding if the number of categories exceeded 5. When working with tree techniques, there is no need for standardization.

\begin{table*}
  \caption{A summary of the key characteristics of the datasets used for evaluation, with $N$, $N_a$ and $N_l$ representing the number of records, attributes and lattice features, respectively. The count of attributes and lattice features was determined after the encoding of categorical attributes. $N_m$ denotes the number of minority samples in classification problems. The columns $par_{t}$ and $par_{f}$ encode the hyperparameters of the trees and forests leading to the highest AUC score ($auc_*$) or $r^2$ score ($r^2_*$). The hyperparameters indicate the minimum number of samples on the leaf nodes ($\alpha$) or the maximum depth ($\beta$), respectively. The datasets are sorted by their size $N\cdot N_a$.}
  \label{tab:datasets}
  \begin{center}
    \begin{scriptsize}
      \begin{tabular}{l@{\hspace{6pt}}r@{\hspace{6pt}}r@{\hspace{6pt}}r@{\hspace{6pt}}r@{\hspace{6pt}}l@{\hspace{6pt}}r@{\hspace{6pt}}l@{\hspace{6pt}}r@{\hspace{6pt}}l@{\hspace{6pt}}r@{\hspace{6pt}}r@{\hspace{6pt}}r@{\hspace{6pt}}l@{\hspace{6pt}}r@{\hspace{6pt}}l@{\hspace{6pt}}r}
\toprule
\multicolumn{9}{c}{Classification} & \multicolumn{8}{c}{Regression} \\
dataset & N & $N_a$ & $N_l$ & $N_m$ & par$_t$ & auc$_t$ & par$_f$ & auc$_f$ & dataset & N & $N_a$ & $N_l$ & par$_t$ & r$^2_t$ & par$_f$ & r$^2_f$ \\
\midrule
appendicitis \cite{keel} & 106 & 7 & 7 & 21 & $\alpha$: 16 & 0.775 & $\alpha$: 16 & 0.852 & diabetes \cite{keel} & 43 & 2 & 2 & $\alpha$: 6 & -0.375 & $\beta$: 2 & -0.083 \\
haberman \cite{keel} & 306 & 3 & 3 & 81 & $\alpha$: 22 & 0.662 & $\alpha$: 13 & 0.721 & o-ring \cite{uci} & 23 & 6 & 4 & $\alpha$: 1 & 0.127 & $\beta$: 2 & 0.150 \\
new-thyroid1 \cite{keel} & 215 & 5 & 5 & 35 & $\alpha$: 15 & 0.959 & $\beta$: 6 & 1.000 & wsn-ale \cite{uci} & 107 & 5 & 4 & $\alpha$: 5 & 0.434 & $\alpha$: 2 & 0.561 \\
glass0 \cite{keel} & 214 & 9 & 9 & 70 & $\alpha$: 15 & 0.837 & $\beta$: 10 & 0.929 & daily-demand \cite{uci} & 60 & 12 & 7 & $\alpha$: 1 & 0.697 & $\beta$: 7 & 0.828 \\
shuttle-6-vs-2-3 \cite{keel} & 230 & 9 & 9 & 10 & $\alpha$: 2 & 1.000 & $\alpha$: 1 & 1.000 & slump-test \cite{krnn} & 103 & 9 & 9 & $\alpha$: 2 & 0.623 & $\beta$: 8 & 0.769 \\
bupa \cite{keel} & 345 & 6 & 6 & 145 & $\alpha$: 11 & 0.695 & $\alpha$: 2 & 0.767 & servo \cite{uci} & 167 & 10 & 2 & $\alpha$: 4 & 0.686 & $\alpha$: 2 & 0.721 \\
cleveland-0-vs-4 \cite{keel} & 177 & 13 & 10 & 13 & $\alpha$: 7 & 0.898 & $\alpha$: 4 & 0.979 & yacht-hydrodynamics \cite{krnn} & 307 & 6 & 6 & $\alpha$: 1 & 0.993 & $\beta$: 11 & 0.995 \\
ecoli1 \cite{keel} & 336 & 7 & 5 & 77 & $\alpha$: 31 & 0.954 & $\alpha$: 3 & 0.958 & autoMPG6 \cite{keel} & 392 & 5 & 5 & $\alpha$: 10 & 0.823 & $\beta$: 12 & 0.872 \\
poker-9-vs-7 \cite{keel} & 244 & 10 & 10 & 8 & $\alpha$: 17 & 0.682 & $\beta$: 4 & 0.986 & excitation-current \cite{uci} & 557 & 4 & 4 & $\beta$: 10 & 1.000 & $\beta$: 11 & 1.000 \\
monk-2 \cite{keel} & 432 & 6 & 4 & 204 & $\alpha$: 2 & 1.000 & $\alpha$: 2 & 1.000 & real-estate-valuation \cite{uci} & 414 & 6 & 5 & $\alpha$: 6 & 0.658 & $\alpha$: 4 & 0.709 \\
hepatitis \cite{krnn} & 155 & 19 & 6 & 32 & $\alpha$: 6 & 0.744 & $\beta$: 2 & 0.871 & wankara \cite{keel} & 321 & 9 & 9 & $\alpha$: 2 & 0.972 & $\beta$: 9 & 0.987 \\
yeast-0-3-5-9-vs-7-8 \cite{keel} & 506 & 8 & 6 & 50 & $\alpha$: 9 & 0.737 & $\alpha$: 4 & 0.810 & plastic \cite{keel} & 1650 & 2 & 2 & $\alpha$: 9 & 0.766 & $\alpha$: 9 & 0.789 \\
mammographic \cite{keel} & 830 & 5 & 5 & 403 & $\alpha$: 46 & 0.900 & $\beta$: 3 & 0.912 & laser \cite{keel} & 993 & 4 & 4 & $\alpha$: 3 & 0.922 & $\beta$: $\infty$ & 0.963 \\
saheart \cite{keel} & 462 & 9 & 8 & 160 & $\alpha$: 25 & 0.719 & $\beta$: 3 & 0.755 & qsar-aquatic-toxicity \cite{uci} & 546 & 8 & 8 & $\alpha$: 17 & 0.355 & $\beta$: 14 & 0.525 \\
page-blocks-1-3-vs-4 \cite{keel} & 472 & 10 & 10 & 28 & $\alpha$: 15 & 0.976 & $\beta$: 12 & 0.999 & baseball \cite{keel} & 337 & 16 & 12 & $\alpha$: 8 & 0.641 & $\beta$: 5 & 0.680 \\
pima \cite{keel} & 768 & 8 & 8 & 268 & $\alpha$: 33 & 0.805 & $\beta$: 5 & 0.836 & maternal-health-risk \cite{uci} & 1013 & 6 & 6 & $\beta$: $\infty$ & 0.713 & $\beta$: $\infty$ & 0.749 \\
wisconsin \cite{keel} & 683 & 9 & 9 & 239 & $\alpha$: 13 & 0.983 & $\alpha$: 2 & 0.993 & cpu-performance \cite{krnn} & 209 & 35 & 6 & $\beta$: 8 & 0.812 & $\beta$: 13 & 0.870 \\
abalone9-18 \cite{keel} & 731 & 9 & 7 & 42 & $\alpha$: 116 & 0.758 & $\alpha$: 4 & 0.850 & airfoil \cite{krnn} & 1503 & 5 & 4 & $\beta$: 15 & 0.861 & $\beta$: $\infty$ & 0.934 \\
winequality-red-3-vs-5 \cite{keel} & 691 & 11 & 11 & 10 & $\alpha$: 85 & 0.815 & $\alpha$: 17 & 0.870 & medical-cost \cite{mlwithr} & 1338 & 6 & 2 & $\beta$: 3 & 0.761 & $\beta$: 3 & 0.765 \\
\bottomrule
\end{tabular}

    \end{scriptsize}
  \end{center}
\end{table*}

\subsection{A note on the use of cross-validation}
\label{sec:tests-cv}

In subsection \ref{sec:hyper}, we discussed why cross-validation cannot be employed to select the conditioning operator as part of the model selection process. Nevertheless, with the right interpretation, cross-validation remains a valuable tool for the quantitative assessment of the bias and the proposed methods: we can estimate the size of the bias, and we can determine which conditioning operator works better given that 80\% of the data is used for training (in a 5-fold scheme). What we cannot do is claim that if one conditioning performs better on a random 80\% of the data, then this remains the case when using all available data for training. Nevertheless, if the proposed method works according to the expectations in relation to the use of the individual operators in the cross-validated evaluation, it is reasonable to assume that this behavior is maintained when all data is used for training.

\subsection{Model selection}
\label{sec:tests-model}

For realistic evaluations, selecting appropriate hyperparameters for trees and forests is crucial. For instance, when building trees with a depth of 1, detecting the effect of the conditioning bias is unlikely. Conversely, growing trees to their maximum depth might result in exaggerated estimations of the bias; however, these models are highly overfitted, leading to poor prediction results.

The objective of model selection is to identify hyperparameters that maximize the performance with the default conditioning operator ($\leq$) of the \emph{sklearn} \cite{sklearn} implementation. Model selection was carried out individually for each dataset using 20 times repeated 5-fold cross-validation and maximizing the ROC AUC score \cite{auc} for classification and the $r^2$ score for regression. The hyperparameters considered were the \emph{minimum number of samples on the leaf nodes} (ranging from 0.5\% to 20\% of the dataset size) and the \emph{maximum depth} (ranging from 2 to 15). The results of the model selection, presenting the optimal hyperparameters and the corresponding AUC and $r^2$ scores, are detailed in Table \ref{tab:datasets}. All subsequent results reported in the paper are achieved using these selected hyperparameters.

As anticipated, the regularization of decision trees is more pronounced compared to random forests, with optimal performance achieved using shallower trees indicated by a higher number of elements on leaf nodes ($\alpha$) or a smaller maximum depth ($\beta$). When the trees are extremely shallow (e.g., when $\alpha$ is greater than about 10\% of the dataset size or $\beta \leq 3$), as is often the case in decision tree classification, the impact of the conditioning bias is expected to be less significant.

\subsection{Hypothesis testing for evaluation}
\label{sec:tests-testing}

\begin{figure}[t]
  \begin{center}
\subfigure[\label{fig:dist-clas}]{
\includegraphics[width=0.22\textwidth]{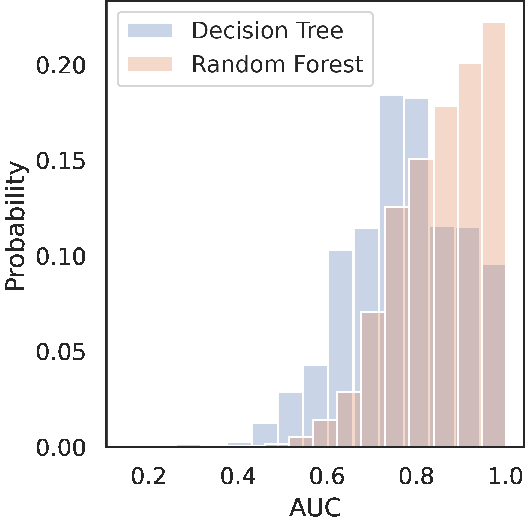}
}
\subfigure[\label{fig:dist-reg}]{
\includegraphics[width=0.22\textwidth]{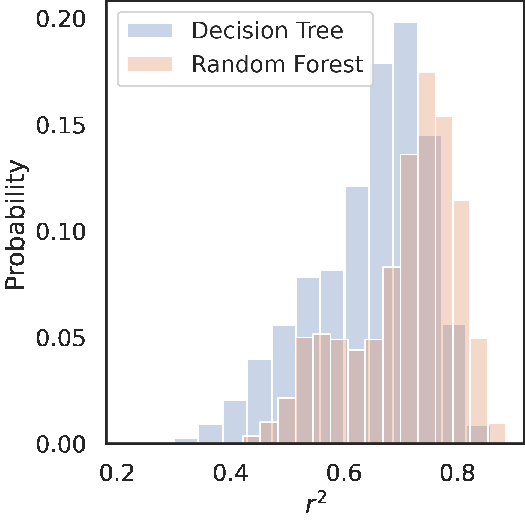}
}
  \end{center}
  \caption{The distributions of the AUC scores in the cross-validation of the \emph{appendicitis} classification dataset (a); and the distributions of $r^2$ scores in the cross-validation of the \emph{real-estate-valuation} regression dataset (b). These distributions are typically skewed, bounded, and eventually non-unimodal, suggesting the preference for the Wilcoxon test over the t-test.}
  \label{fig:dist}
\end{figure}

We are set to investigate a presumably small effect. In addition to the commonly used approach of comparing the mean AUC or $r^2$ scores in cross-validation evaluations, we employ statistical hypothesis testing to distinguish signals from noise. The evaluations are carried out in thorough, 400 times repeated 5-fold cross-validations, using the same folds for a particular dataset and each classifier or regressor. Since the distributions of AUC and $r^2$ scores are not normal (see Figure \ref{fig:dist}), we utilize Wilcoxon's signed-rank test, considered a generalization of the t-test, for the comparison of the locations of distributions.

Consider two variations of a method, denoted as $A$ and $B$ (for example, $A$ using default conditioning, $B$ using non-default conditioning), evaluated on the same folds, resulting in a paired sample of performance scores $(s^{i}_A, s^{i}_B)_{i=1}^{2000}$. The null hypothesis posits that $s^i_A - s^i_B$ has a symmetric distribution around zero, implying $\mathbb{P}(s^i_A > s^i_B) = 1/2$, that is, neither variant tends to consistently outperform the other. We use two alternative hypotheses:
\begin{enumerate}
  \item the distribution of $s^i_A - s^i_B$ is not symmetric about zero, implying that $\mathbb{P}(s^i_A > s^i_B) \neq 1/2$, that is, it is more likely to get a greater score with one of the methods than with the other. This hypothesis is used to test if there is any difference between the methods $A$ and $B$.
  \item the distribution of $s^i_A - s^i_B$ is stochastically greater than a distribution symmetric about zero, implying that $\mathbb{P}(s^i_A > s^i_B) \geq \mathbb{P}(s^i_B > s^i_A)$, that is, it is more likely to get a higher score with the method $A$ than using $B$. This hypothesis is used for the testing that method $A$ improves the performance compared to method $B$.
\end{enumerate}
In the rest of the section, to reduce clutter, we do not report p-values, rather, we report whether the p-value of a test falls below the significance level of 0.05. This signifies the presence of statistical evidence against the null hypothesis, in favor of the alternative hypothesis. Finally, while the Wilcoxon test is related to the locations of the medians rather than the relation of the means, it is observed that on many folds there is no difference in the performance of two variations of an estimator, leading to identical medians. Therefore, as an indicative absolute measure, the mean performance scores and their differences are reported alongside the results of the hypothesis testing.

\subsection{Quantitative illustration of the effect}
\label{sec:tests-quant}

In this subsection, we investigate the existence of the conditioning bias quantitatively, using decision tree and random forest classification and regression.

\begin{table*}[t]
  \caption{The results of investigating the presence of the bias. In each block (classification and regression with decision trees and random forests), the columns $\rho$ and $\rho_k$ indicate the proportion of nodes with thresholds on feature domain values when the model is fitted to all data and 80\% of the data in cross-validation, respectively. The columns $auc_d$ and $r^2_d$ contain the differences of mean performance scores of using the default ($\leq$) and non-default ($<$) operators for inference. Finally, in the columns $p_{\neq}$, the symbol $\ast$ indicates that statistical evidence is found against the equality of the distributions, that is, the p-value of the Wilcoxon test is less than 0.05.}
  \label{tab:proof}
  \begin{center}
    \begin{small}
      \begin{tabular}{l@{\hspace{4pt}}r@{\hspace{4pt}}r@{\hspace{4pt}}r@{\hspace{4pt}}c@{\hspace{4pt}}r@{\hspace{4pt}}r@{\hspace{4pt}}r@{\hspace{4pt}}c@{\hspace{4pt}}l@{\hspace{4pt}}r@{\hspace{4pt}}r@{\hspace{4pt}}r@{\hspace{4pt}}c@{\hspace{4pt}}r@{\hspace{4pt}}r@{\hspace{4pt}}r@{\hspace{4pt}}c}
\toprule
\multicolumn{9}{c}{Classification} & \multicolumn{9}{c}{Regression} \\
dataset & \multicolumn{4}{c}{Decision Tree} & \multicolumn{4}{c}{Random Forest} & dataset & \multicolumn{4}{c}{Decision Tree} & \multicolumn{4}{c}{Random Forest} \\
 & $\rho$ & $\rho_{k}$ & auc$_{d}$ & p$_{\neq}$ & $\rho$ & $\rho_{k}$ & auc$_{d}$ & p$_{\neq}$ &  & $\rho$ & $\rho_{k}$ & r$^2_{d}$ & p$_{\neq}$ & $\rho$ & $\rho_{k}$ & r$^2_{d}$ & p$_{\neq}$ \\
\midrule
appendicitis & 0 & .00 &  0 &  & .02 & .03 & -4e-06 &  & diabetes & 0 & .02 & -7e-05 &  & 0 & .00 &  7e-04 & $\ast$ \\
haberman & 0 & .02 & -3e-04 & $\ast$ & .04 & .05 &  6e-06 &  & o-ring & 0 & .03 & -4e-03 & $\ast$ & .13 & .10 & -2e-02 & $\ast$ \\
new-thyroid1 & 0 & .01 &  4e-06 &  & .07 & .09 &  8e-05 & $\ast$ & stock-portfolio & .05 & .03 & -1e-04 &  & .02 & .03 &  2e-05 & $\ast$ \\
glass0 & 0 & .03 &  0 &  & .05 & .05 & -2e-05 &  & wsn-ale & 0 & .12 & -1e-03 &  & .12 & .11 & -1e-03 & $\ast$ \\
shuttle-6-vs-2-3 & 0 & 0 &  0 &  & .14 & .13 &  0 &  & daily-demand & 0 & .01 & -3e-04 & $\ast$ & .03 & .03 & -7e-04 & $\ast$ \\
bupa & .12 & .10 &  1e-04 & $\ast$ & .25 & .24 &  2e-03 & $\ast$ & slump-test & .10 & .09 &  5e-04 & $\ast$ & .18 & .16 & -4e-04 & $\ast$ \\
cleveland-0-vs-4 & 0 & .01 & -3e-03 &  & .09 & .11 & -8e-04 & $\ast$ & servo & 0 & 0 & -7e-06 &  & .00 & .00 & -9e-06 & $\ast$ \\
ecoli1 & 0 & .06 &  0 &  & .12 & .12 & -7e-05 & $\ast$ & yacht-hydrodynamics & 0 & 0 &  0 &  & .00 & .00 &  0 &  \\
poker-9-vs-7 & 0 & 0 &  0 &  & .14 & .16 & -2e-03 & $\ast$ & autoMPG6 & .09 & .05 & -3e-04 & $\ast$ & .18 & .17 &  3e-05 &  \\
monk-2 & 0 & 0 &  0 &  & .00 & .00 &  0 &  & excitation-current & .10 & .10 & -4e-08 & $\ast$ & .13 & .12 &  2e-07 & $\ast$ \\
hepatitis & 0 & .04 & -2e-04 &  & .03 & .04 & -3e-04 & $\ast$ & real-estate-valuation & .02 & .02 & -4e-04 & $\ast$ & .03 & .03 & -2e-06 &  \\
yeast-0-3-5-9-vs-7-8 & 0 & .04 & -4e-05 &  & .09 & .09 & -2e-04 & $\ast$ & wankara & .02 & .02 & -5e-05 & $\ast$ & .04 & .04 & -3e-05 & $\ast$ \\
mammographic & 0 & .00 &  3e-05 & $\ast$ & .02 & .04 &  6e-07 & $\ast$ & plastic & 0 & .02 &  1e-03 & $\ast$ & .02 & .02 &  7e-04 & $\ast$ \\
saheart & 0 & .03 & -8e-06 & $\ast$ & .07 & .08 &  6e-05 & $\ast$ & laser & .25 & .28 & -9e-04 & $\ast$ & .38 & .38 & -2e-04 & $\ast$ \\
page-blocks-1-3-vs-4 & .25 & .12 &  1e-04 & $\ast$ & .06 & .06 & -4e-06 &  & qsar-aquatic-toxicity & 0 & .00 &  0 &  & .03 & .03 & -3e-04 & $\ast$ \\
lymphography & 0 & 0 & -4e-03 &  & .04 & .04 & -1e-03 & $\ast$ & baseball & .09 & .14 & -1e-03 &  & .23 & .24 &  7e-04 & $\ast$ \\
pima & 0 & .05 & -3e-04 & $\ast$ & .16 & .19 &  4e-06 &  & maternal-health-risk & .22 & .23 &  1e-03 & $\ast$ & .24 & .23 & -7e-04 & $\ast$ \\
wisconsin & 0 & .03 & -1e-04 & $\ast$ & .09 & .12 &  4e-05 &  & cpu-performance & .23 & .21 &  2e-03 & $\ast$ & .24 & .22 &  2e-03 & $\ast$ \\
abalone9-18 & 0 & 0 &  0 &  & .08 & .08 & -1e-06 &  & airfoil & .00 & .01 &  9e-06 & $\ast$ & .01 & .01 & -1e-05 & $\ast$ \\
winequality-red-3-vs-5 & 0 & 0 &  0 &  & .07 & .06 & -4e-04 & $\ast$ & medical-cost & 0 & .00 &  0 &  & .00 & .00 &  2e-06 & $\ast$ \\
\bottomrule
\end{tabular}

    \end{small}
  \end{center}
\end{table*}

First, we investigate whether thresholds truly fall on domain values. To achieve this, we fitted trees and forests with the default conditioning operator and determined the proportion of nodes with thresholds falling on observed feature values in two scenarios: during the 400 times repeated 5-fold cross-validation ($\rho_k$) and separately, by fitting the models to all available data ($\rho$). The results are summarized in the corresponding columns of Table \ref{tab:proof}. Overall, the cross-validation does not significantly exaggerate the effect, as there are only slight differences in the proportions when fitted to all data ($\rho$) compared to only 80\% of the data in the cross-validation iterations ($\rho_k$). The only exception is classification with decision trees, where fitting to all data eliminates biasing nodes in most cases. This can be explained by considering that decision tree classification is the most strongly regularized problem in model selection, leading to trees with only 10-15 nodes in most cases. Nevertheless, the results confirm the existence of \emph{nodes with thresholds falling on feature domain values}, even when the models are fit to all available data.

Second, we investigate whether the choice of conditioning could have a statistically significant effect on inference. The cross-validated evaluation was carried out with the non-default conditioning operator, and the results were compared in terms of the differences in average AUC and $r^2$ scores ($auc_d$ and $r^2_d$), the score of the non-default conditioning subtracted from that of the default one. We also used the Wilcoxon test to see if evidence against the stochastic equality of the distributions ($p_{\neq}$) can be found. As the results summarized in Table \ref{tab:proof} show, \emph{the choice of the conditioning operator can a have statistically significant effect on performance}. In the strongly regularized case of decision tree classification, only about 30\% of the datasets are affected. However, in the other scenarios, the majority of the cases show an impact. Regarding the differences in mean performance scores, the variation can reach up to $0.3$ percentage points of AUC score in classification (\emph{cleveland-0-vs-4} dataset with decision tree) and 2 percentage points of $r^2$ in regression (\emph{o-ring} dataset with random forest). 

Overall, the results are aligned with the expectations: (a) there can be nodes with thresholds falling on feature domain values; (b) the choice of conditioning can have a detectable and statistically significant effect on the performance. Finally, we highlight that the direction of the effect depends on both the problem and the modeling technique: in some cases the default, in other cases the non-default conditioning leads to better performance, proving the point that the prior choice in popular implementations can be detrimental in certain cases.

\subsection{Evaluation of the proposed method}
\label{sec:tests-proposed}

As established in Section \ref{sec:proposed}, \emph{the primary goal of the proposed methods is to eliminate the risk of using a conditioning that has a detrimental effect on the performance} (which might eventually be the conditioning supported by an implementation by default). The tests conducted in this subsection are tailored to assess this specific aspect of the problem.

\begin{table*}[t]
  \caption{The results of the evaluation. The columns $p_{\neq}$ are identical to those in Table \ref{tab:proof}, indicating if statistical evidence was found against the equivalence of the two conditionings. The columns $p_{\leq}$ (or $p_{<}$) indicate if the proposed method stochastically dominates (+) or minorizes (-) the use of the conditioning operator $\leq$ (or $<$) with statistical significance. The columns $auc_d$ and $r^2_d$ contain the difference in the mean performance score of the proposed method and that of the worst performing conditioning, indicating the improvement over the worst case. To reduce clutter, we did not populate the cells where the proposed method cannot be distinguished with statistical significance from any of the individual conditionings.}
  \label{tab:results}
  \begin{center}
    \begin{small}
      \begin{tabular}{l@{\hspace{4pt}}l@{\hspace{4pt}}l@{\hspace{4pt}}l@{\hspace{4pt}}l@{\hspace{8pt}}l@{\hspace{4pt}}l@{\hspace{4pt}}l@{\hspace{4pt}}l@{\hspace{8pt}}l@{\hspace{4pt}}l@{\hspace{4pt}}l@{\hspace{4pt}}l@{\hspace{4pt}}l@{\hspace{8pt}}l@{\hspace{4pt}}l@{\hspace{4pt}}l@{\hspace{4pt}}l}
\toprule
\multicolumn{9}{c}{Classification} & \multicolumn{9}{c}{Regression} \\
dataset & \multicolumn{4}{c}{Decision Tree} & \multicolumn{4}{c}{Random Forest} & dataset & \multicolumn{4}{c}{Decision Tree} & \multicolumn{4}{c}{Random Forest} \\
 & p$_{\neq}$ & p$_{\leq}$ & p$_{<}$ & auc$_{d}$ & p$_{\neq}$ & p$_{\leq}$ & p$_{<}$ & auc$_{d}$ &  & p$_{\neq}$ & p$_{\leq}$ & p$_{<}$ & r$^2_{d}$ & p$_{\neq}$ & p$_{\leq}$ & p$_{<}$ & r$^2_{d}$ \\
\midrule
appendicitis &  &  &  &  &  &  &  &  & diabetes &  &  &  &  & $\ast$ & - & + & 3.3e-04 \\
haberman & $\ast$ & + & - & 2.2e-04 &  &  &  &  & o-ring & $\ast$ & + & - & 2.2e-03 & $\ast$ & + & - & 1.5e-02 \\
new-thyroid1 &  &  &  &  & $\ast$ &  &  &  & stock-portfolio &  &  &  &  & $\ast$ & - & + & 1.7e-05 \\
glass0 &  &  &  &  &  &  &  &  & wsn-ale &  &  &  &  & $\ast$ & + & - & 5.8e-04 \\
shuttle-6-vs-2-3 &  &  &  &  &  &  &  &  & daily-demand & $\ast$ & + & - & 2.3e-04 & $\ast$ & + & - & 3.5e-04 \\
bupa & $\ast$ &  & + & 4.1e-04 & $\ast$ & - & + & 1.2e-03 & slump-test & $\ast$ & + & - & 1.0e-03 & $\ast$ & + & - & 3.0e-04 \\
cleveland-0-vs-4 &  &  &  &  & $\ast$ & + & - & 3.5e-04 & servo &  &  &  &  & $\ast$ & + & - & 4.3e-06 \\
ecoli1 &  &  &  &  & $\ast$ & + & - & 3.0e-05 & yacht-hydrodynamics &  &  &  &  &  &  &  &  \\
poker-9-vs-7 &  &  &  &  & $\ast$ & + & - & 9.8e-04 & autoMPG6 & $\ast$ & + &  & 2.9e-04 &  & + & + & 6.9e-05 \\
monk-2 &  &  &  &  &  &  &  &  & excitation-current & $\ast$ & + & + & 2.6e-07 & $\ast$ &  & + & 2.1e-07 \\
hepatitis &  &  &  &  & $\ast$ & + &  & 1.5e-04 & real-estate-valuation & $\ast$ & + &  & 3.2e-04 &  &  & + & 8.8e-06 \\
yeast-0-3-5-9-vs-7-8 &  &  &  &  & $\ast$ & + & - & 1.3e-04 & wankara & $\ast$ & + &  & 5.5e-05 & $\ast$ & + & - & 1.7e-05 \\
mammographic & $\ast$ & - & + & 1.3e-05 & $\ast$ & + &  & 1.4e-05 & plastic & $\ast$ & - & + & 7.7e-04 & $\ast$ & - & + & 3.6e-04 \\
saheart & $\ast$ &  & + & 6.2e-05 & $\ast$ & - &  & 2.8e-05 & laser & $\ast$ & + & + & 1.6e-03 & $\ast$ & + & + & 2.2e-04 \\
page-blocks-1-3-vs-4 & $\ast$ &  & + & 1.5e-04 &  &  &  &  & qsar-aquatic-toxicity &  &  &  &  & $\ast$ & + & - & 1.9e-04 \\
lymphography &  &  &  &  & $\ast$ & + &  & 1.1e-03 & baseball &  & + & + & 2.2e-03 & $\ast$ & - & + & 5.3e-04 \\
pima & $\ast$ & + & - & 2.1e-04 &  &  &  &  & maternal-health-risk & $\ast$ & - & + & 1.5e-03 & $\ast$ & + & - & 5.5e-04 \\
wisconsin & $\ast$ & + & - & 7.9e-05 &  &  &  &  & cpu-performance & $\ast$ &  & + & 5.6e-03 & $\ast$ & - & + & 1.6e-03 \\
abalone9-18 &  &  &  &  &  &  &  &  & airfoil & $\ast$ & - & + & 2.1e-05 & $\ast$ & + & - & 5.1e-06 \\
winequality-red-3-vs-5 &  &  &  &  & $\ast$ & + & - & 2.1e-04 & medical-cost &  &  &  &  & $\ast$ & - &  & 1.1e-06 \\
\bottomrule
\end{tabular}

    \end{small}
  \end{center}
\end{table*}

The proposed methods were evaluated in the same 400 times repeated 5-fold cross-validation experiments and on the same folds we used in the previous section. Wilcoxon's one-sided tests were employed to compare the performance with that obtained by the two conditioning operators. The expectation is that the proposed method outperforms the worst performing conditioning although it might be outperformed by the better conditioning. 

The results summarized in Table \ref{tab:results} can be interpreted as follows. Similar to Table \ref{tab:proof}, $p_{\neq}$ indicates if statistically significant evidence was found against the equivalence of the conditionings. The columns $p_{\leq}$ (and $p_{<}$) indicate if statistical evidence is found against the equivalence of the proposed method and using the operator $\leq$ (and $<$), in favor of dominating it (+) or minorizing it (-). Empty cells indicate that there is no statistically significant evidence for a difference (the p-value is less than 0.05). Finally, the columns $auc_d$ and $r^2_d$ contain the mean performance improvement of the proposed method compared to the worst performing conditioning. For example, the entry for the \emph{haberman} dataset with decision tree classification reads as follows: there is statistical evidence for the conditioning bias ($p_{\neq}$ is $\ast$); for the proposed method stochastically dominating the conditioning $\leq$ ($p_{\leq}$ is $+$); and for the proposed method being stochastically dominated by the conditioning $<$ ($p_{<}$ is $-$).

The analysis of the results shows that the proposed methods never yield worse results (with statistical significance) than both of the individual conditioning operators. Given that one cannot know which conditioning leads to better results, the results suggest that \emph{there is no more risk of using the proposed methods than using one particular conditioning}. Furthermore, in all but two cases when the two conditionings lead to different results ($p_{\neq} is \ast$), the proposed methods yield better results (with statistical significance) than at least one of the individual conditioning operators, suggesting that \emph{the proposed methods mitigate the risk of performance deterioration caused by using the worse performing (eventually default) conditioning}. In the two exceptions (\emph{saheart} dataset with random forest classification and the \emph{medical-cost} dataset with random forest regression) the performance of the proposed method is equivalent to the worse performing conditioning, with no statistical evidence for difference.

Interestingly, in certain cases of regression, the proposed method outperforms both conditioning operators with statistical significance (the \emph{laser} dataset with both decision tree and random forest, the \emph{baseball} dataset with decision tree, and \emph{autoMPG6} dataset with random forest). This can be explained by the fact that both conditionings might have detrimental effects on different test records, and the averaging by the proposed method can lead to better performance than both.

Finally, we investigate how much improvement the proposed methods achieve compared to the worst-performing conditioning in terms of average performance scores. Again, we focus on those cases when there is a statistically significant difference, otherwise, the variation is considered to be numerical noise. In the case of the highly regularized decision tree classification, only minor differences can be observed, up to 0.02 percentage points of AUC score (dataset \emph{haberman}); with random forest classification the highest improvement is 0.1 percentage points of AUC score (dataset \emph{bupa}); with decision tree regression even 0.5 percentage points improvement is observed in the $r^2$ score (dataset \emph{cpu-performance}); and with random forest regression, improvements up to 1.5 percentage points of $r^2$ score (dataset \emph{o-ring}) are observable. 

The higher improvements in regression compared to classification can be explained by the sensitivities of the performance scores to small changes. The $r^2$ score of regression has a continuous dependence on the predictions, sensitive to the slightest changes. Contrarily, even though the predicted probabilities are adjusted in the right directions, the AUC score of classification does not change until the order of predicted probabilities belonging to at least one pair of positive and negative test samples does not change.

Overall, we can conclude that on the test datasets, the proposed method works according to the expectations: in most cases, when there is a difference between the performance of the two conditionings, the proposed method leads to better results than the worst-performing conditioning, with statistical significance. The stronger regularization of decision trees and the insensitivity of the AUC score to small changes leads to smaller effects in decision trees and classification problems, but in random forest regression, even an improvement of 1.5 percentage points of $r^2$ score has been achieved.

\section{Summary and Conclusions}
\label{sec:summary}

Decision trees and random forests are undoubtedly among the most commonly used supervised learning techniques, and the most popular implementations (such as \cite{sklearn, tree}) are built on binary decision trees. In this paper, we investigated the effect of certain implicit and hidden modeling choices of binary decision tree implementations, such as using the mid-points between observations as threshold ($t$) and conditioning in the form $x\leq t$ or $x < t$, on the performance.

In Section \ref{sec:problem}, we discussed in detail that, although the form of conditioning has a negligible effect on continuous features, it can interfere with the thresholds and impact predictions when features take equidistant values with relatively high probability. (In \ref{sec:tests-datasets}, we found this type of feature to be relatively common in practice, see Table \ref{tab:datasets}.) The variation being caused by the modeling choices, we refer to it as the \emph{conditioning bias}.

In Section \ref{sec:proposed}, we proposed the elimination of the bias by integrating out the dependence on the conditioning, which can yield better performance than using the implicit, default choice of a particular implementation. In subsection \ref{sec:equiv}, we discussed how the form of conditioning is related to the mirroring of trees and building trees from the opposites (additive inverses) of feature vectors. Based on these observations, in subsection \ref{sec:integrate}, we have proposed multiple ways to perform the averaging, supporting various scenarios depending on the accessibility of fitted decision trees. As an important finding, with random forests, the elimination of the conditioning bias incurs no additional computational costs in training and prediction time.

In Section \ref{sec:tests}, we have conducted extensive experiments with decision trees and random forests using 400 times repeated 5-fold cross-validation, incorporating 20 classification and 20 regression datasets. In subsection \ref{sec:tests-quant}, we demonstrated the detectability of the conditioning bias with statistical significance, revealing variations of up to 0.1-0.2 percentage points of AUC and $r^2$ scores (see Table \ref{tab:proof}). From the results we conclude that \emph{the conditioning bias can affect the prediction results in applications where features with lattice characteristics are present}.
In subsection \ref{sec:tests-proposed}, we demonstrated that in the vast majority of cases the proposed method leads to better performance than the worst performing conditioning operator, with statistical significance. In terms of absolute figures, the smallest improvements were observed with the strongly regularized decision tree classification up to 0.02 percentage points of AUC score, and the highest improvements are achieved with random forest regression, up to 1.5 percentage points of $r^2$ scores (see Table \ref{tab:results}). We conclude that \emph{the proposed methods are effective techniques to reduce the risk of deteriorated performance due to using the implicitly supported but underperforming conditioning}.

Finally, we mention that the implementation of the entire analysis is available at the GitHub repository \url{https://github.com/gykovacs/conditioning_bias}.

\section*{Acknowledgements}

This work was developed within the scope of the project
i3N, UIDB/50025/2020 and UIDP/50025/2020, financed by
national funds through the FCT/MEC-Portuguese Foundation for Science and Technology. G.T. was supported by FCT
Grant No. CEECIND/03838/2017.

\appendix
\section{Sample data}
\label{sec:sample}

The decision tree in Figure \ref{fig:dt} can be inferred from the following training records representing a loan evaluation problem: 
\begin{center}
  \begin{footnotesize}
    \begin{tabular}{rrrr}
      \toprule
      self-employed & income (k) & dependents & target \\
      \midrule
      0 & 80 & 1 & 1 \\
      0 & 60 & 3 & 1 \\
      0 & 80 & 1 & 0 \\
      1 & 50 & 2 & 0 \\
      1 & 50 & 4 & 1 \\
      1 & 70 & 4 & 0 \\
      \bottomrule
    \end{tabular}
  \end{footnotesize}
\end{center}
The target label encodes if the applicant will default (1) or pay back (0). The tree is induced by the \emph{sklearn} (version 1.3.2) \cite{sklearn} decision tree classifier using the random seed 5.

\section*{Declaration of generative AI and AI-assisted technologies in the writing process}

During the preparation of this work the author(s) used ChatGPT 3.5 in order to improve the use of the English language. After using this tool/service, the author(s) reviewed and edited the content as needed and take(s) full responsibility for the content of the publication.

\bibliographystyle{model2-names}
\bibliography{refs}

\end{document}